\documentclass[onefignum,onetabnum]{siamart190516}

\usepackage{svg}
\usepackage{bbm}
\usepackage{lipsum}
\usepackage{amsfonts}
\usepackage{graphicx}
\usepackage{epstopdf}
\usepackage{algorithmic}
\usepackage{nomencl}
\makenomenclature

\ifpdf
  \DeclareGraphicsExtensions{.eps,.pdf,.png,.jpg}
\else
  \DeclareGraphicsExtensions{.eps}
\fi

\newsiamremark{remark}{Remark}
\newsiamremark{hypothesis}{Hypothesis}
\crefname{hypothesis}{Hypothesis}{Hypotheses}
\newsiamthm{claim}{Claim}

\headers{Multiresolution Convolutional Autoencoders}{Y. Liu, C. Ponce, S. L. Brunton, and J. N. Kutz}

\title{Multiresolution Convolutional Autoencoders \thanks{Submitted to the editors DATE.}}

\author{Yuying Liu\thanks{Department of Applied Mathematics, University of Washington, Seattle, WA 
  (\email{yliu814@uw.edu}).}
\and Colin Ponce\thanks{Lawrence Livermore National Lab, Livermore, CA}
\and Steven L. Brunton \thanks{Mechanical Engineering, University of Washington, Seattle, WA} \and J. Nathan Kutz \thanks{Department of Applied Mathematics, University of Washington, Seattle, WA}}

\usepackage{amsopn}

\ifpdf
\hypersetup{
  pdftitle={Multiresolution Convolutional Autoencoders},
  pdfauthor={Y. Liu, C. Ponce, S. L. Brunton, and J. N. Kutz}
}
\fi

\begin{document}

\maketitle

\begin{abstract}
We propose a {\em multi-resolution convolutional autoencoder} (MrCAE) architecture that integrates and leverages three highly successful mathematical architectures: (i) multigrid methods, (ii) convolutional autoencoders and (iii) transfer learning.  The method provides an adaptive, hierarchical architecture that capitalizes on a progressive training approach for multiscale spatio-temporal data.   This framework allows for inputs across multiple scales: starting from a compact (small number of weights) network architecture and low-resolution data, our network progressively deepens and widens itself in a principled manner to encode new information in the higher resolution data based on its current performance of reconstruction. Basic transfer learning techniques are applied to ensure information learned from previous training steps can be rapidly transferred to the larger network. As a result, the network can dynamically capture different scaled features at different depths of the network. The performance gains of this adaptive multiscale architecture are illustrated through a sequence of numerical experiments on synthetic examples and real-world spatial-temporal data. 
%
\end{abstract}

\begin{keywords}
  convolutional autoencoder, multiresolution analysis, multigrid, transfer learning, model scaling, multi-scale dynamics
\end{keywords}

\begin{AMS}
  65T99, 37N10, 37M10
\end{AMS}

\section{Introduction}
\label{sec:intro}
The multiscale spatio-temporal dynamics observed in many complex systems poses significant challenges for modeling and prediction.  Although we are often primarily interested in macroscale phenomena, the microscale dynamics must also be modeled and understood, as it plays an important role in driving the macroscale behavior. Indeed, a given system may have multiple fast and slow time scales as well as a number of micro to macro spatial scales that interact to produce the complex dynamics observed.  This makes modeling multiscale systems particularly difficult unless the time scales are disambiguated in a principled way.  Coarse graining and mesh-refinement (multigrid methods) are two principled mathematical techniques for addressing multiscale behavior.  In the former, the microscale physics are averaged over to produce an effective macroscale model, while in the latter, computational models are refined where the macroscale variables produce large errors.  In this paper, we present a {\em multi-resolution convolutional autoencoder} (MrCAE), a decomposition scheme inspired by multigrid computational methods to progressively refine a multiscale description of spatio-temporal data.  It leverages and exploits aspects of multigrid methods and transfer learning to produce an effective multi-scale analysis tool for characterizing large scale spatio-temporal data.

Multigrid methods~\cite{mccormick1987multigrid,trottenberg2000multigrid} have been extensively developed for physics-based simulation models where coarse grained models must be progressively refined in order to achieve a required numerical precision while keeping the simulation tractable.  Multigrid architectures provide a principled method for targeting the refinement process, constituting a mature field with wide spread applications in the engineering and physical sciences.  In contrast, coarse graining methods attempt to construct a macroscale physics model by progressive construction of coarse grained variables and their dynamics.  Mathematical algorithms such as the {\em  heterogeneous multiscale modeling} (HMM)~\cite{weinan_heterognous_2003,weinan_principles_2011} and {\em equation-free method}~\cite{kevrekidis_equation-free_2003} provide principled methods for multiscale systems.
Additional work has focused on testing for the presence of multiscale dynamics so that analyzing and simulating multiscale systems is more computationally efficient \cite{froyland_computational_2014,froyland_trajectory-free_2016}.

Data-driven methods, specifically neural networks (NNs), have emerged as an attractive alternative for characterizing multi-scale physics~\cite{gonzalez1998identification,yang2018physics,Wehmeyer2018jcp,Mardt2018natcomm,lusch2018deep,champion2019data,he2019mgnet,Raissi2019jcp,raissi2020science}.   The structure of {\em convolutional neural networks} (CNNs) are especially relevant for multiscale data as the convolutional window extracts features of the data at an appropriate, coarse- or fine-grained resolution.  As a result, CNNs have demonstrated exceptional performance in image processing tasks~\cite{goodfellow2016deep}.  Indeed, in the wake of AlexNet~\cite{krizhevsky2012imagenet}, many aspects of CNN design have been thoroughly studied individually, such as the spatial filters~\cite{szegedy2015going, simonyan2014very}, nonlinear activation functions~\cite{xu2015empirical}, width and depth of the network~\cite{zagoruyko2016wide}, skip connections~\cite{He_2016_CVPR}, batch normalization~\cite{ioffe2015batch}, etc~\cite{alom2018history}. As more complex neural architectures and designs arise, NNs have become increasingly popular, which enables the process of automating architecture engineering by jointly considering all design factors~\cite{tan2019efficientnet, liu2018progressive, zoph2016neural}. However, the design choices of NN algorithms are highly automated and often uninterpretable, which makes integrating domain knowledge difficult.



Although CNNs can capitalize on the multiscale features of data, because they begin processing with only small local patches, they often fail to exploit the large-scale, low-dimensional structure typically observed in spatio-temporal data.  Such structure is routinely leveraged for reduced order modeling~\cite{benner2015survey,kutz2013data,brunton2019data,quarteroni2014reduced,hesthaven2016certified,Taira2017aiaa} of high-dimensional spatio-temporal systems.  This motivates the innovations of the current work; here, we propose a multi-resolution convolutional autoencoder (MrCAE) that begins by processing on downsampled (ie ``coarse'') data in order to capture large-scale, low-dimensional structure, and then progressively refines both the data and the neural network while employing transfer learning in building each stage's neural network. This progress-refinement approach provides a principled framework for leveraging multiresolution  and transfer learning ideas, enabling one to  build multiscale models for spatio-temporal data  that result in more compact networks and more compact encodings than those produced by traditional CAE methods. We envision our proposed architecture to be a critical piece of many multi-scale neural network architectures, especially as encoder-decoder based models have already resulted in many successful applications, such as linearization of dynamics \cite{lusch2018deep, gin2019deep}, model discovery via sparse regression \cite{champion2019data}, forecasting \cite{rudy2019deep}, etc. 

Importantly, our architecture leverages data in an intelligent way:  querying data that targets the refinement process.  Thus a hierarchy of models is constructed with less data, producing highly compact networks and encodings. The method is well aligned with the arguments in \cite{chen2015net2net}: \textit{'Initially, a small model may be preferred, in order to prevent overfitting and to reduce the computational cost of using the model. Later, a large model may be necessary to fully utilize the large dataset.'}.

Our paper is organized as follows: the multi-resolution network architecture is proposed in \cref{sec:arch}, the progressive training algorithm is presented in \cref{sec:train}, experimental results are in \cref{sec:experiments}, and the conclusions and discussions follow in \cref{sec:discussions}. Our code is publicly available at \url{https://github.com/luckystarufo/MrCAE}.

\section{Network Architecture}
\begin{figure}[t]
    \centering
    \label{fig:arch1}\includegraphics[width=130mm, height=54mm]{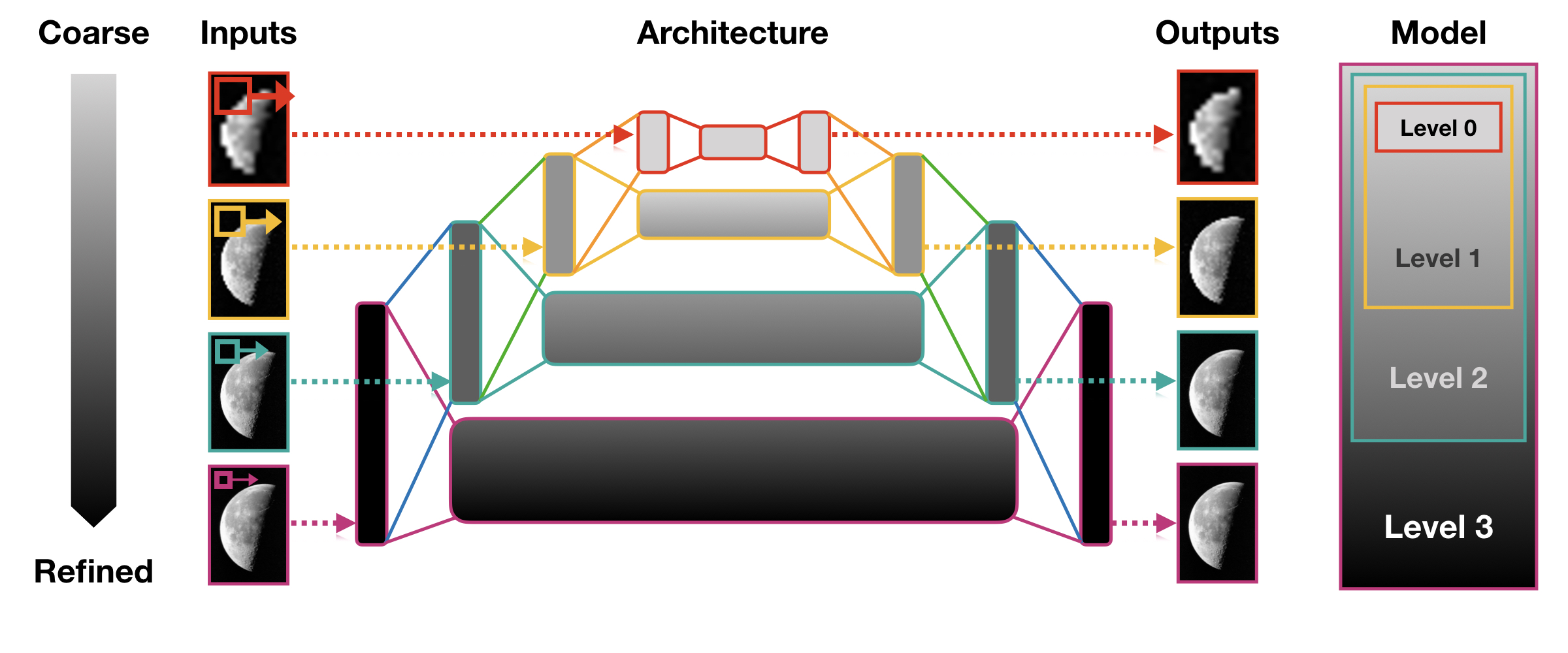}
    \caption{\textbf{A Schematic Overview of the Network Architecture}. In this example, there are $4$ levels for the architecture which are colored in red, yellow, cyan and purple respectively. They are recursively built up to process data across different resolutions --- architectures built for processing coarser data are later embedded into the next-level architectures to ensure knowledge transfer. }
\end{figure}

    \cref{fig:arch1} details the overall architecture of our MrCAE. It consists of several levels of autoencoders. Each level is associated with the data of a particular resolution. The network is built up recursively: in the base level, a small autoencoder is trained to encode and reconstruct the coarsest data. We then refine the data so that our inputs are of a higher resolution and embed the existing architecture into a new autoencoder (the transfer learning step) to learn the new, refined data. In this way, we construct a hierarchy of trained neural networks. The mathematical details of the construction are given in the subsections that follow. 
    \par

\label{sec:arch}
\begin{figure}[t]
    \centering
    \label{fig:arch2}\includegraphics[width=130mm, height=61mm]{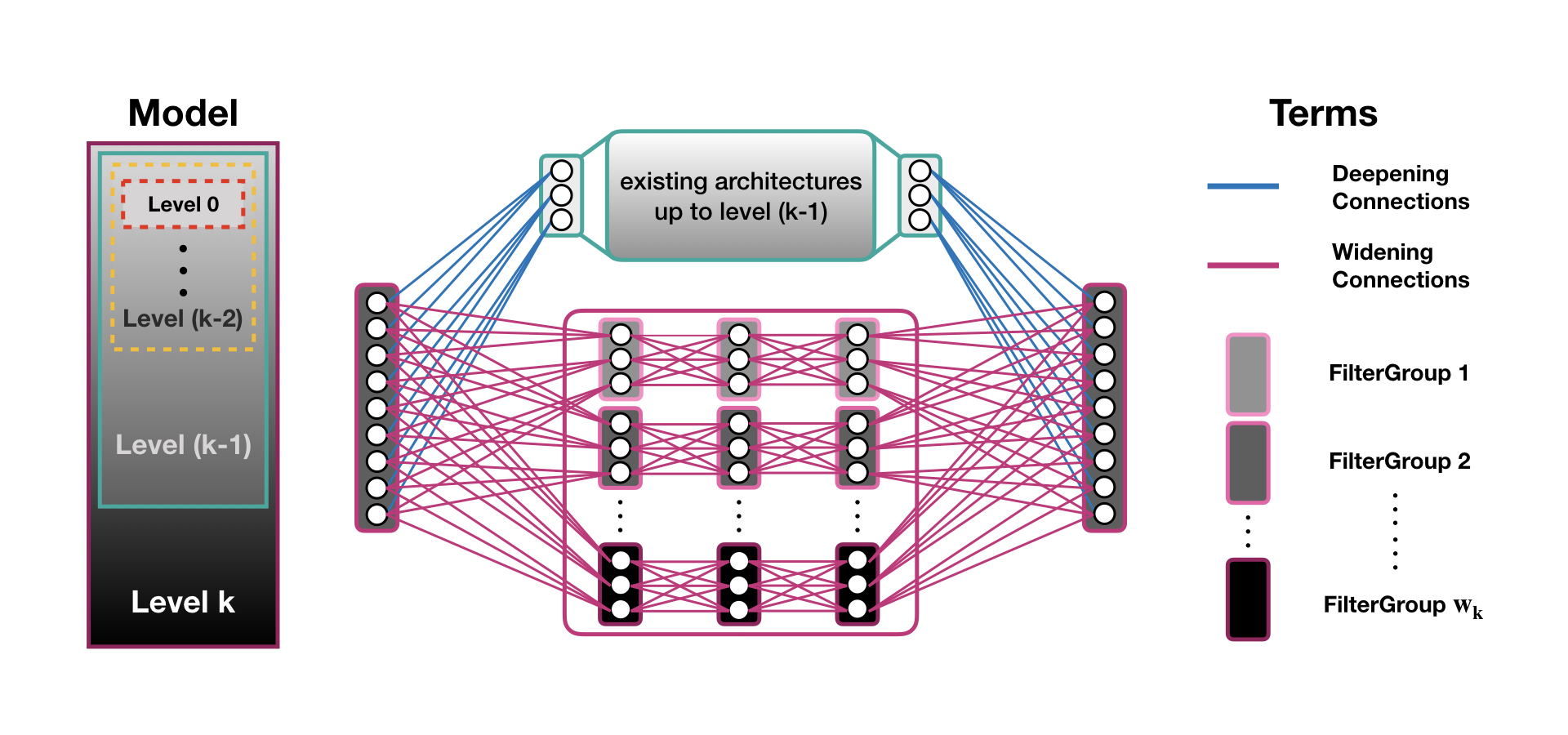}
    \caption{\textbf{Build up Network Architecture at Level $k$}. Within level $k$, we perform one deepening operation and a sequence of widening operations. The deepening operation (shown in blue) is the transfer learning step, creating a convolutional filter that connects the new input (fine) to the previous level input (coarse). Widening operations (shown in purple) are performed sequentially by allocating more convolutional filters so that it can capture new, higher-resolution features.}
\end{figure}
\begin{figure}[t]
    \centering
    \label{fig:arch3}\includegraphics[width=130mm, height=73mm]{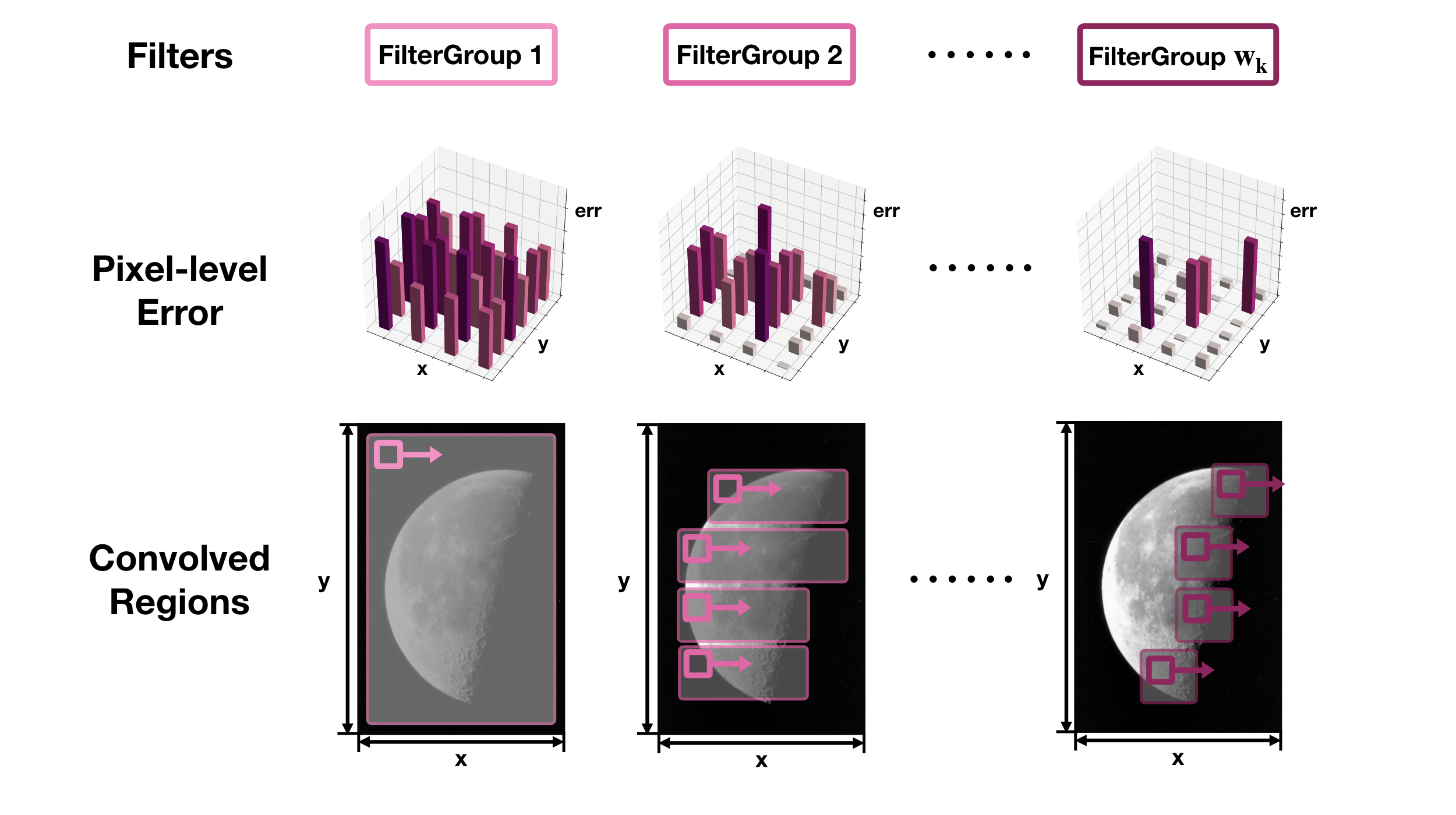}
    \caption{\textbf{Adaptive Filters (I)}. For each widening operation shown in \cref{fig:arch2}, new convolutional filters are only applied to the regions that are still poorly resolved. The progressive refinement ensures a parsimonious use of the parameters.}
\end{figure}

\subsection{Hierarchical Construction}
    The technical details for constructing each level of the network is demonstrated in \cref{fig:arch2}: we perform a deepening operation (except for the base level) followed by a sequence of widening operations.
    \par
    The \emph{deepening operation} inserts a convolutional/deconvolutional layer between the current and previous level inputs/outputs. This is the transfer learning step. We denote the inserted convolutional filter as $c^{(k)}$ and the deconvolutional filter as $d^{(k)}$, and let $f_{\boldsymbol{\theta_{k-1}}}$ be the existing network at level $(k-1)$, $g^{(k)}$ be the network after we apply this deepening operation. Mathematically, we have
    \begin{equation}
        g^{(k)} = d^{(k)} \circ f_{\boldsymbol{\theta_{k-1}}} \circ c^{(k)}
    \end{equation}
    In our setup, we make the resolution differ by a factor of $2$ along each spatial dimension between two adjacent levels, which is typical of a multi-resolution analysis~\cite{kutz2013data}. As a result, convolutional/deconvolutional filters with a kernel size $3 \times 3$ and stride size $2$ are used to ensure compatible dimensions. Similar to \cite{chen2015net2net}, the two inserted layers are properly initialized to ensure knowledge transfer. More technical details about the initialization are covered in the \cref{sec:filter_init}. There is another, perhaps more intuitive, way to understand the design: data at coarser levels can be regarded as obtained by applying a sequence of hard-coded down-sampling (convolutional) operations to the finest data and vice versa. When we proceed to a new level, the network replaces one of these hard-coded operators with trainable convolutional/deconvolutional filters.  
    \par
    The \emph{widening operation} expands the network capacity in order to capture the new, finer-grained features of the higher resolution data. Specifically, it adds a group of new convolutional pathways through the network, and then continues training. See the FilterGroups in Figure \ref{fig:arch2}. Note that each new pathway is separate, not adding any connections to old pathways; this reduces the risks of overfitting. Mathematically speaking, the widening operations at level $k$ create a list of new pathways for the network, we denote them as $w_1^{(k)}, w_2^{(k)}, \cdots, w_{k_n}^{(k)}$. Then we have,
    \begin{equation}
        f_{\boldsymbol{\theta_{k}}} = g^{(k)} + w_1^{(k)} + \cdots + w_{w_k}^{(k)}.
    \end{equation}
    
\subsection{Adaptive Filters}
    The way we perform a sequence of widening operations is shown in \cref{fig:arch3}. Suppose we have performed $j$ groups of widening operations and finished the subsequent training. We proceed by computing the mean squared error of a $3 \times 3$ region around each pixel. Our $(j+1)^{th}$ group of filters will be only applied to the regions where the reconstructions exhibit high error; details are explained in \cref{sec:train}. Then a subsequent training is performed for the new architecture. We do this recursively until all regions can be reconstructed within some error threshold/bound. This process bears a close resemblance to mesh refinement \cite{berger1989local} of multigrid methods~\cite{mccormick1987multigrid,trottenberg2000multigrid} and therefore is highly adaptive and ensures a parsimonious use of parameters. 
    \par
    It is worth noting that we do not use nonlinear activation functions, in contrast to the classical designs in which convolutional filters are usually followed by a Rectified Linear Units (ReLU). We find it actually gives us a performance boost in terms of the training progress while also making the filters highly interpretable. In our design, the highly adaptive filters play the role of the nonlinear activation functions: by explicitly specifying where to apply the filters, we are forcing those filters to be activated in the relevant regions while leaving them inhibited in other regions. Additionally, this implies that the resulting, trained network is fully linear, enabling extremely fast matrix-based representations.

\section{Training}
\label{sec:train}
\subsection{Framework}
To accompany our adaptive architecture, we have developed a progressive training framework which is outlined in \cref{alg:pt}. The training is performed whenever the architecture changes, thus it can effectively utilize the new pathways for representing new data. Early stopping criteria are implemented as well, however, we reference the full training details in \cref{sec:early_stop} in order to maintain clarity and simplicity for the description of the architecture.
\begin{algorithm}
\caption{Progressive Training}
\label{alg:pt}
\begin{algorithmic}
\STATE{Define $N$ to be the number of levels}
\STATE{Define $E$ to be the maximum training epochs}
\STATE{Define training data with increasing resolutions $\{D^{(0)}, \cdots, D^{(N-1)}\}$}
\STATE{Initialize a model object $M$}
\FOR{i in $0, 1, \cdots, N-1$:}
    \STATE{Perform deepening operation on $M$;}
    \STATE{Perform $E$ epochs training;}
    \WHILE{Reconstruction at this level is not fully resolved}
        \STATE{Perform widening operation on $M$;}
        \STATE{Perform $E$ epochs training;}
    \ENDWHILE
\ENDFOR
\RETURN $M$
\end{algorithmic}
\end{algorithm}

\subsection{Loss Function}
Our loss function is designed to capture the general, low-rank dynamics as well as identifying outlier/singularities. It is of the same functional form across each level.  However, different resolution data are used to calculate this quantity. Suppose we are training the network at level $k$ ($0 \leq k < N$) with data set $D^{(k)} \in R^{m_k \times n_k \times T_k}$ ($m_k$, $n_k$, $T_k$ are the width, height of each snapshot and the number of snapshots), and $\widehat{D}^{(k)} := f_{\boldsymbol{\theta_k}}(D^{(k)})$ is the corresponding reconstruction through the network, where $\boldsymbol{\theta_k}$ are the parameters of the current network $f_{\boldsymbol{\theta_k}}$ at level $k$\footnote{In fact, at level $k$, the network keeps growing itself as we perform the deepening operation and the sequence of widening operations, which gives rise to different networks $g_k$, $f_{\boldsymbol{\theta_k}}^{(1)}, \cdots, f_{\boldsymbol{\theta_k}}^{(k_n)}$ respectively. Here, without causing ambiguity, we refer them all as $f_{\boldsymbol{\theta_k}}$.}. Let $i$, $j$, $t$ to be the indices of the row, column and snapshot. We formulate our loss function $\mathcal{L}(\boldsymbol{\theta_k}, D^{(k)})$ to be the following:
\begin{equation}
    \label{eq:loss}
    \mathcal{L}(\boldsymbol{\theta_k}, D^{(k)}) = \omega \mathcal{L}_{mse}(\boldsymbol{\theta_k}, D^{(k)}) + (1-\omega)\mathcal{L}_{max}(\boldsymbol{\theta_k}, D^{(k)})
\end{equation}
where
\begin{equation}
    \label{eq:mseloss}
    \mathcal{L}_{mse}(\boldsymbol{\theta_k}, D^{(k)}) = \frac{1}{m_k n_k T_k} \sum_{i=1}^{m_k}\sum_{j=1}^{n_k}\sum_{t=1}^{T_k} (D^{(k)}_{i,j,t} - \widehat{D}^{(k)}_{i,j,t})^2
\end{equation}
\begin{equation}
    \label{eq:maxloss}
    \mathcal{L}_{max}(\boldsymbol{\theta_k}, D^{(k)}) = \max_{i,j}\frac{1}{T_k}\sum_{t=1}^{T_k}(D^{(k)}_{i,j,t} - \widehat{D}^{(k)}_{i,j,t})^2
\end{equation}
The first term in \cref{eq:loss} is the classic mean squared error (MSE) to ensure an overall satisfactory reconstruction.  The second term captures the worst case scenario which primarily corresponds to the highly variable dynamics over some regions in the spatio-temporal data. The loss function is a weighted sum of these two terms modulated by a control parameter $\omega$ ($0 \leq \omega \leq 1$). Based on our experience, we find a small $\omega$ (eg. $0.5$) is ideal for dynamics that have a clear separation of spatial scales, so that the network pays more attention to the singularities or highly variable regions. But for those without persistent spatial patterns, we recommend setting $\omega$ close to $1$.

\subsection{Measuring Error}
\label{sec:err}
In addition to the loss function in \cref{eq:loss}, we define two other important quantities. 
The first quantity is developed for tracking the overall training progress. The loss function in \cref{eq:loss} only gives us the information of how it performs at a specific level. Nothing can be said about the reconstructions at higher levels. In this case, knowledge transfer between adjacent levels cannot be evaluated. What we need is a metric that reflects the overall training progress. Therefore, we define a metric that estimates the reconstruction error with respect to the finest resolution data at each level:
\begin{equation}
    \label{eq:global_losses}
    \mathcal{L}^{global}(\boldsymbol{\theta}, D^{(N-1)}) = \omega\mathcal{L}_{mse}^{global}(\boldsymbol{\theta}, D^{(N-1)}) + (1-\omega)\mathcal{L}_{max}^{global}(\boldsymbol{\theta}, D^{(N-1)})
\end{equation}
\begin{equation}
    \mathcal{L}_{mse}^{global} = \frac{1}{m_{N-1} n_{N-1} T_{N-1}} \sum_{i=1}^{m_{N-1}}\sum_{j=1}^{n_{N-1}}\sum_{t=1}^{T_{N-1}} (D^{(N-1)}_{i,j,t} - \mathcal{U}^{(N-k-1)}(\widehat{D}^{(k)}_{i,j,t}))^2
\end{equation}
\begin{equation}
    \mathcal{L}_{max}^{global} = \max_{i, j}\frac{1}{T_{N-1}}\sum_{t=1}^{T_{N-1}} (D^{(N-1)}_{i,j,t} - \mathcal{U}^{(N-k-1)}(\widehat{D}^{(k)}_{i,j,t}))^2
\end{equation}
where 
\begin{equation}
    \mathcal{U}^{(s)} = \underbrace{\mathcal{U} \circ \mathcal{U} \circ \cdots \circ \mathcal{U}}_{s} 
\end{equation}
is a composition of a sequence of up-sampling operation $\mathcal{U}$ which is chosen to be a bi-linear interpolation with a kernel size $3 \times 3$ and a stride size $2$. This is similar to the prolongation operator from multigrid methods. It should be noted that this metric is only proposed for the purpose of validating the effectiveness of knowledge transfer across different levels of the model, it would be absent in the realistic settings where we have a growing data set so that the finest data $D^{(N-1)}$ may not be accessible initially. 
\par
\begin{figure}[t]
    \centering
    \label{fig:filters}\includegraphics[width=130mm, height=60mm]{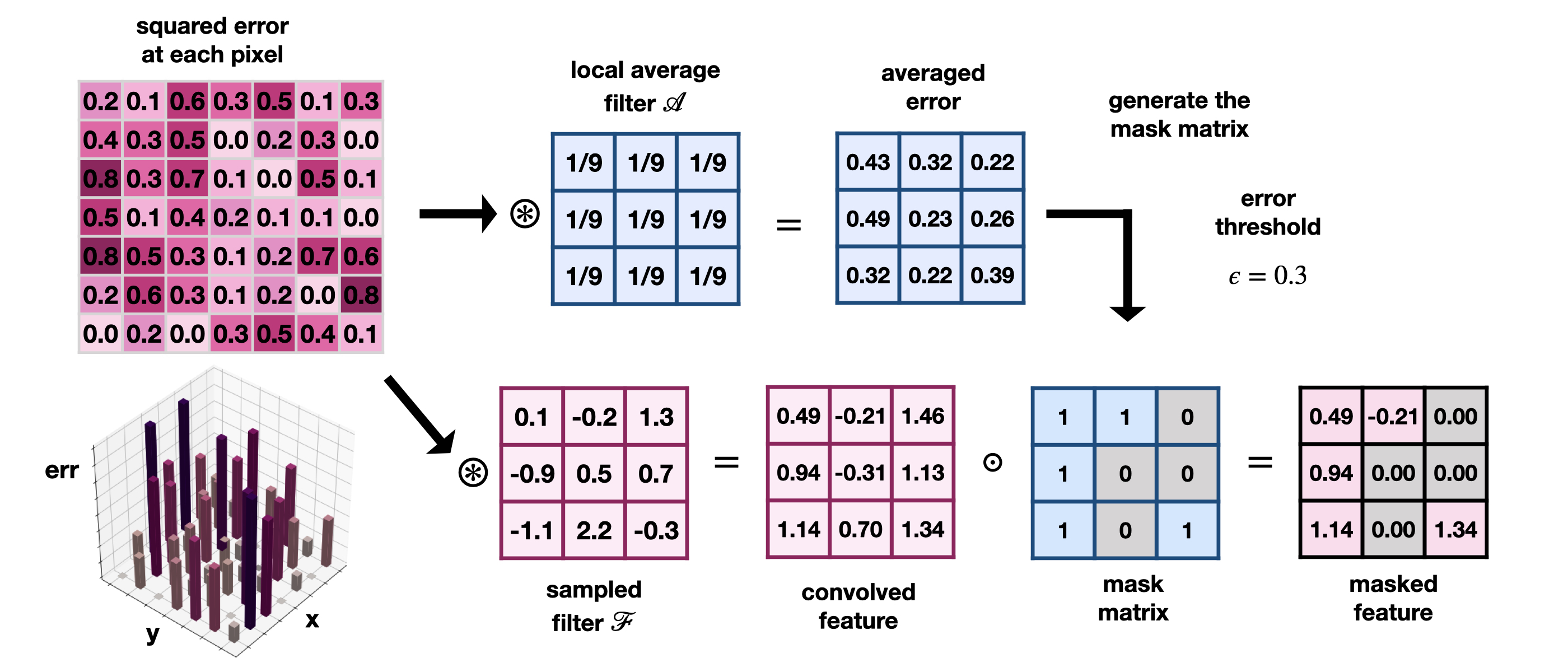}
    \caption{\textbf{Adaptive Filters (II)}. This picture illustrates the implementation details of applying adaptive filters. It is done by calculating the mask  (shown at the top) and convolved features of all regions (shown at the bottom) followed by a point-wise multiplication.}
\end{figure}
The other metric is used to provide feedback for the adaptive filters. Within a widening operation, the new group of convolutional filters are only applied to the regions that still need to be further refined. Technically, this is achieved by applying the filters to all regions of the inputs first (and therefore obtaining the convolved features), then determining the irrelevant ones. To tell if a convolved feature should be masked out or not, we first calculate the mean squared error of each pixel along the time axis, followed by performing a down-sampling operation $\mathcal{A}$ (local average) since the width and height of the convolved features are both reduced by a factor of $2$ compared to the original inputs. Finally, we threshold it by a prescribed tolerance $\epsilon$ to obtain the mask $\mathcal{M}$:
\begin{equation}
    \mathcal{M} = \mathbbm{1}\bigg\{\mathcal{A}\Big[\frac{1}{T_k}\sum_{t=1}^{T_k}(D^{(k)}_{\boldsymbol{:},\boldsymbol{:},t} - \widehat{D}^{(k)}_{\boldsymbol{:},\boldsymbol{:},t})^2\Big]_{i, j} \geq \epsilon\bigg\}
\end{equation}
The process is illustrated in \cref{fig:filters}.

\section{Experimental results}
\label{sec:experiments}
In this section, we test the performance of our MrCAE on a range of  data displaying increasing levels of complexity. 
Several unsteady fluid flow fields are analyzed, as fluid dynamics exhibit a range of multiscale behavior and have been the focus of intense modeling efforts, with and without machine learning~\cite{Taira2017aiaa,Brunton2020arfm}.  
In the first part, we show the performance of our network on each individual case.  Specifically, we show the reconstructions on some testing snapshots and heat maps of the regions being refined across each level and the error plot with respect to \cref{eq:global_losses} that reflects the overall training progress. In the second part, we benchmark our network against a structurally similar network: residual encoder-decoder network (RED-Net\cite{peng2018red}). We show our architecture scales better in terms of the number of network parameters and the size of encoding.
\par
All data are resized so that each snapshot has the shape $(2^p-1) \times (2^q-1)$ (with some positive integers $p$ and $q$) before data ingestion into the network,  all snapshots are shuffled randomly and split in to training set ($70\%$), validation set ($20\%$) and testing set ($10\%$).

\subsection{Individual experiments}

\begin{figure}[t]
    \centering
    \label{fig:toy1}\includegraphics[width=130mm, height=150mm]{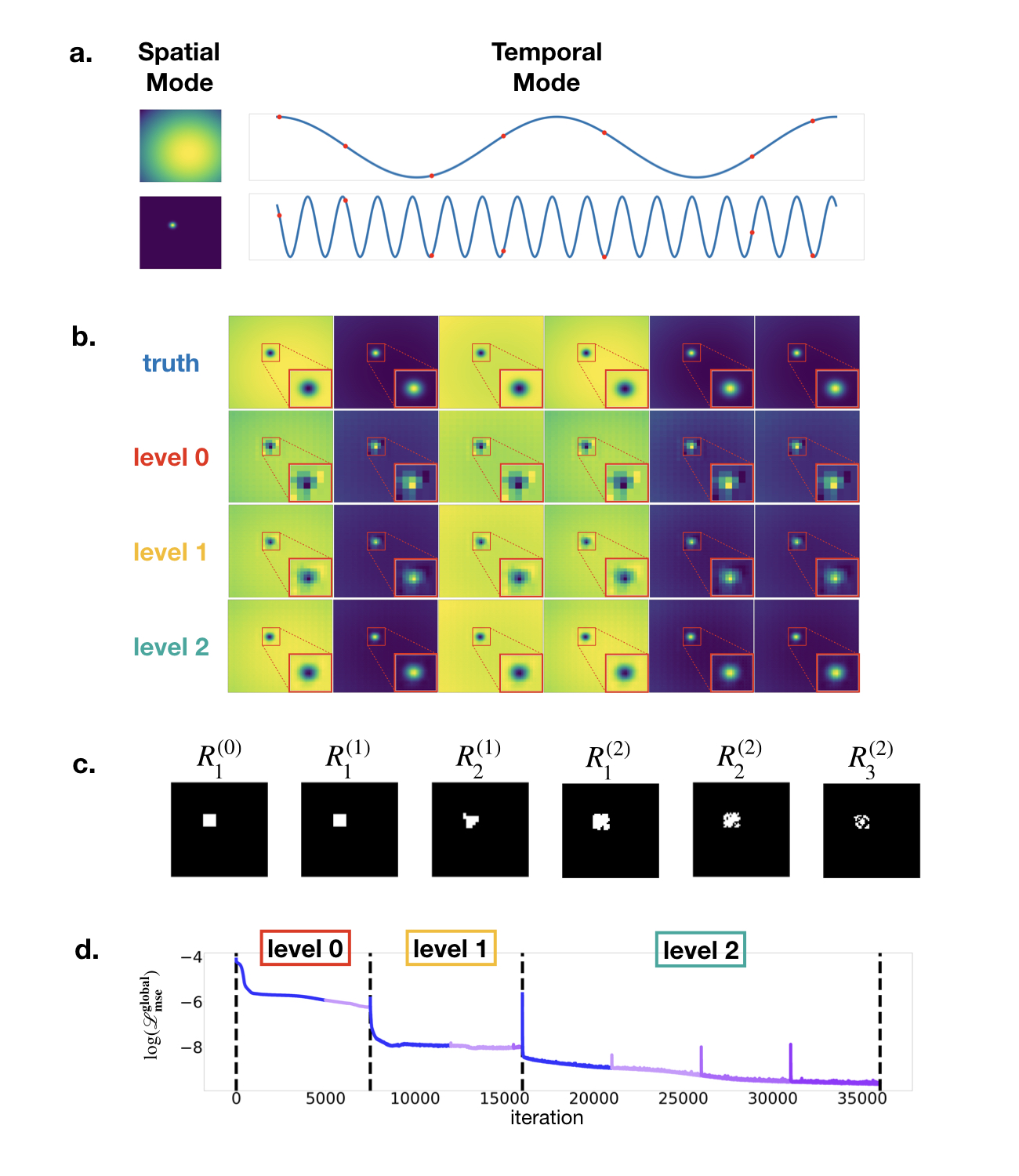}
    \vspace*{-.5in}
    \caption{\textbf{a}. Two separated spatial and temporal modes for the example of two oscillatory modes. Red dots correspond to the sampled test snapshots. \textbf{b}. Original data and the reconstructions of the sampled test snapshots across each level of the network. \textbf{c}. Regions that different groups of adaptive filters apply across different levels of the architecture. Here, $R^{(k)}_j$ represents the regions that the $j^{th}$ group of adaptive filters being applied at level $k$. \textbf{d}. Logarithmic error plot (in terms of the metric presented in \cref{sec:err}) on the validation set throughout the training.}
\end{figure}

\subsubsection{Two oscillatory modes}
For the first example (\cref{fig:toy1}), we consider two nonlinear spatial modes driven by sinusoidal temporal modes with different frequencies:
\begin{equation}
    \Phi(x, y, t) = u(x, y)\cos{(\omega_0 t)} + v(x, y)\cos{(\omega_1 t + \frac{\pi}{4})}
\end{equation}
where
\begin{equation}
    \begin{split}
        u(x, y) &= \cosh{(\frac{x+1}{\sigma_0})}\cosh{(\frac{y-1}{\sigma_0})} \\
        v(x, y) &= \frac{1}{(2\pi\sigma_1)^{1/2}}\exp{(-\frac{(x-1)^2+(y+1)^2}{2\sigma_1^2})}
    \end{split}
\end{equation}
We generate it in the domain of $[-5, 5] \times [-5, 5]$ using $127 \times 127$ grids, and $500$ snapshots are uniformly collected from the time interval $[0, 8\pi]$. We set $\omega_0 = 0.5$, $\omega_1 = 4.0$, $\sigma_0 = 10.0$ and $\sigma_1 = 0.25$ in our experiment.

\cref{fig:toy1}a shows the synthetic spatio-temporal data. We visualize the two different spatial modes and their associated temporal modes. $6$ sampled snapshots are drawn from the test set which are marked in red. This is an example where small spatial (high-frequency) content is modulated by a large, slow-varying background mode. 
\par
For this example, we construct a network of $3$ levels, with each level having $1$, $2$ and $3$ groups of adaptive filters (a.k.a. widening operations) correspondingly. The output reconstructions across different levels shown in \cref{fig:toy1}b become sharper and sharper as the network grows. The regions that adaptive filters being applied are shown in \cref{fig:toy1}c. One can clearly see the network faithfully picks up the high-frequency content with more convolutional filters while leaving other regions less parametrized which is consistent with our intuition. We can also see the proposed error metric on the validation set keeps decreasing throughout the training process except for the starting phase of each operation, which is the effects of random initialization, suggesting an effective knowledge transfer.

\subsubsection{Two oscillatory modes with one drifting}
For the second example, we also consider two nonlinear modes driven by different sinusoidal temporal modes. However, only one of the nonlinear modes is purely spatial, the other one is modulated by time. This synthetic example is meant to replicate the effects of traveling waves or the drifting dynamics that usually appears in the spatial-temporal data. Mathematically, we have:
\begin{equation}
    \Phi(x, y, t) = u(x)\cos{(\omega_0 t)} + v(x, t)\cos{(\omega_1 t + \frac{\pi}{4})}
\end{equation}
where
\begin{equation}
    \begin{split}
        u(x, y) &= \cosh{(\frac{x+1}{\sigma_0})}\cosh{(\frac{y-1}{\sigma_0})} \\
        v(x, y, t) &= \frac{1}{(2\pi\sigma_1)^{1/2}}\exp{(-\frac{(x-3+0.5t)^2+(y+3-0.5t)^2}{2\sigma_1^2})}
    \end{split}
\end{equation}

\begin{figure}[t]
    \centering
    \vspace{-.1in}
    \label{fig:toy2}\includegraphics[width=130mm, height=172mm]{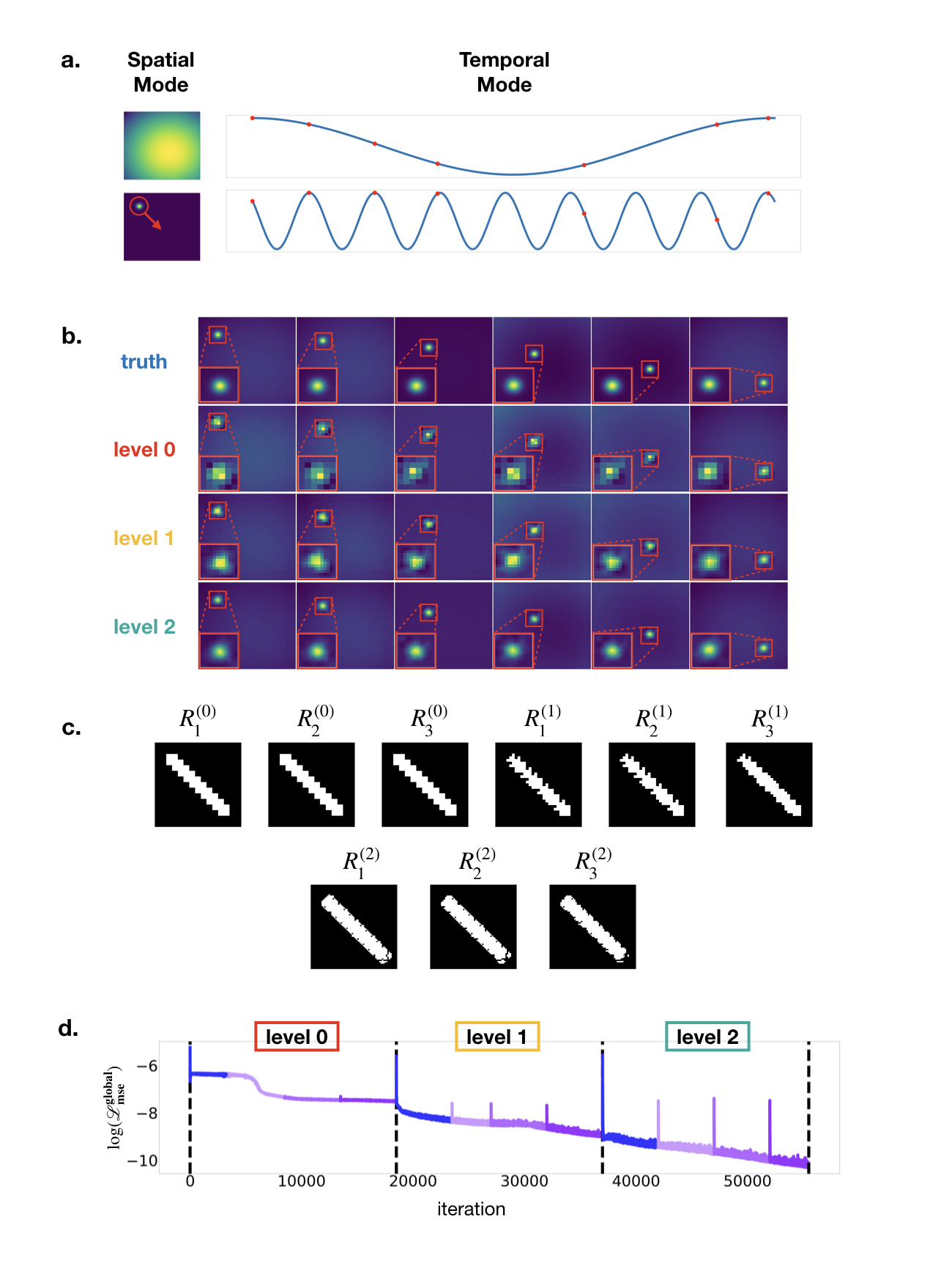}
    \vspace*{-.5in}
    \caption{From top to bottom: \textbf{a}. Two separated spatial and temporal modes for the example of two oscillatory modes with one drifting. Red dots correspond to the sampled test snapshots. \textbf{b}. Original data and the reconstructions of the sampled test snapshots across each level of the network. \textbf{c}. Regions that different groups of adaptive filters apply across different levels of the architecture. Here, $R^{(k)}_j$ represents the regions that the $j^{th}$ group of adaptive filters being applied at level $k$. \textbf{d}. Logarithmic error plot (in terms of the metric presented in \cref{sec:err}) on the validation set throughout the training.}
\end{figure}

We generate the data in the domain of $[-5, 5] \times [-5, 5]$ using $127 \times 127$ grids, and $500$ snapshots are uniformly collected from the time interval $[0, 4\pi]$. We set $\omega_0 = 0.5$, $\omega_1 = 4.0$, $\sigma_0 = 10.0$, $\sigma_1 = 0.25$ and $v=1.0$ in our experiment. \cref{fig:toy2} shows the two spatial-temporal modes, sampled test snapshots, reconstructions across different levels, refined regions and the logarithmic error plot.

In this example, we construct a network of $3$ levels, with $3$, $3$ and $3$ groups of adaptive filters applied on each level. Similarly, we see the network keeps refining itself by growing new pathways, our adaptive filters successfully capture the high-frequency contents drifting along the diagonal and knowledge learned in the previous level can be successfully transferred to the next level.

\subsubsection{Channel flow}
The channel flow dataset is acquired from the Johns Hopkins Turbulence Database \cite{li2008public, perlman2007data, graham2016web}. Data was generated by solving the Navier-Stokes equations in a domain of size $8\pi \times 2 \times 3\pi$ using $2048 \times 512 \times 1536$ nodes. For more details, refer to \cite{kanov2015johns}. We sample the slice of $z=1.5\pi$ and down-sample the other two spatial dimensions both by a factor of $8$, so that the spatial dimensions of our snapshots are $255 \times 63$. $500$ snapshots are collected with $\Delta t=0.052$ between each snapshot.


\begin{figure}[t]
    \vspace*{-.1in}
    \centering
    \label{fig:channel_flow}\includegraphics[width=130mm]{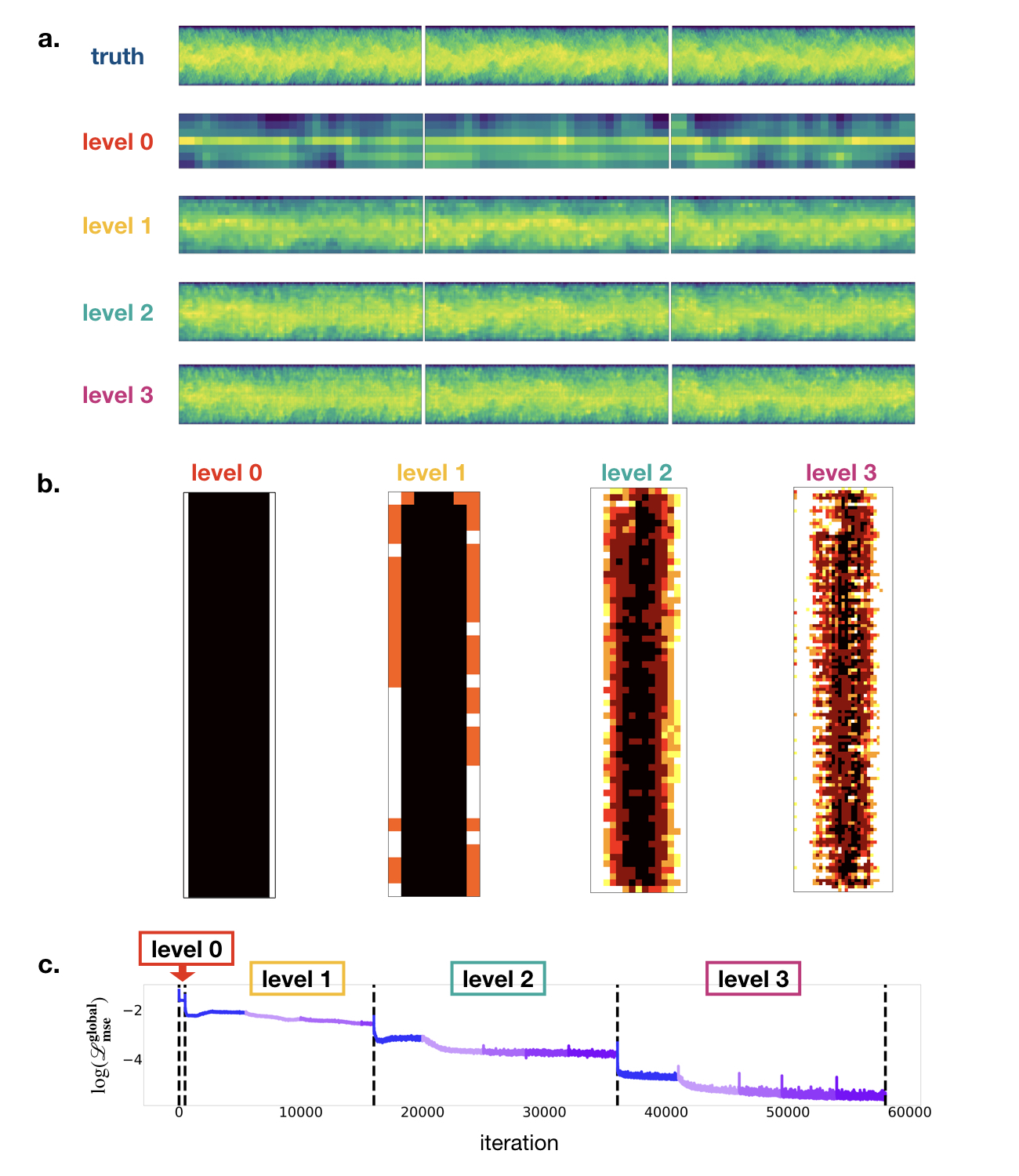}
    \vspace*{-.4in}
    \caption{\textbf{a}. Original data and reconstructions of the channel flow example across different levels of the network over the sampled test snapshots. \textbf{b}. Heat maps that reflect the regions that adaptive filters apply across different levels of the architecture. The brighter the pixel, the more filters are applied. \textbf{c}. Logarithmic error plot (in terms of the metric presented in \cref{sec:err}) on the validation set throughout the training.}
\end{figure}

In this example, we set up a network of $4$ levels with each level having $0$, $3$, $3$, $4$ groups of adaptive filters. (At level $0$, no adaptive filters are used because all pixels can be well-reconstructed with the deepening operation alone.) \cref{fig:channel_flow} shows the increasingly refined outputs at each level and the heat maps of regions that the adaptive filters apply. Notably, we see from the heat maps that the network spends more effort processing the regions near the boundary which intuitively makes sense: there are more detailed small-scale vortical behaviors arising within these regions. The plot of the decreasing error on the validation set also justifies the knowledge transfer of each operation.

\subsubsection{Forced isotropic turbulence}
The forced isotropic turbulence data set is also acquired from the Johns Hopkins Turbulence Database \cite{li2008public, perlman2007data, yeung2012dissipation}. The data is generated from a direct numerical simulation of forced isotropic turbulence on a $1024 \times 1024 \times 1024$ periodic grid, using a pseudo-spectral parallel code. The range of spatial dimensions x, y and z are $[0, 2\pi]$.  We sample the slice of $z=\pi$ and down-sample the other two spatial dimensions both by a factor of $8$ so that each snapshot is of size $127 \times 127$. $503$ snapshots are collected with $\Delta t=0.02$ between each snapshot. This is an example without a clear separation of scales, although it still possesses rich microscopic dynamics across all regions.
\par

\begin{figure}[t]
    \centering
    \label{fig:fluid}\includegraphics[width=130mm, height=150mm]{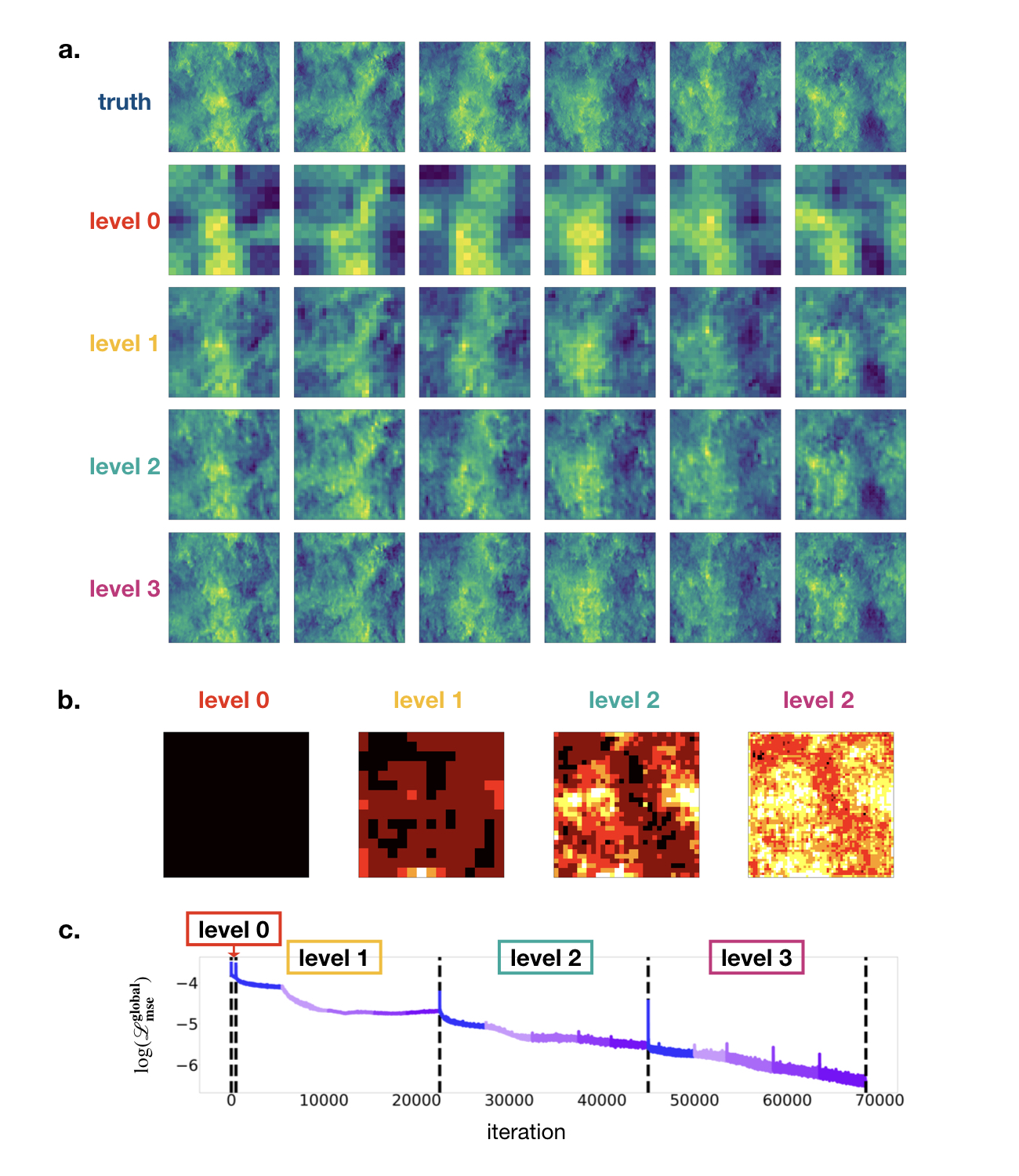}
    \vspace*{-.4in}
    \caption{\textbf{a}. Original data and the reconstructions of the forced isotropic turbulence example across different levels of the network over the sampled test snapshots. \textbf{b}. Heat maps that reflect the regions that adaptive filters apply across different levels of the architecture. The brighter the pixel, the more filters are used. \textbf{c}. Logarithmic error plot (in terms of the metric presented in \cref{sec:err}) on the validation set throughout the training.}
\end{figure}

For this example, we set up a network of $4$ levels with each level having $0$, $4$, $4$, $4$ groups of adaptive filters. (At level $0$, no adaptive filters are used because all pixels can be well-reconstructed with the deepening operation alone.) As shown in \cref{fig:fluid}, the network still offers reasonably well and progressively refined reconstructions of sampled test snapshots and the error plot also shows the effectiveness in knowledge transfer. But the heat maps of refined regions do not exhibit clear spatial patterns due to the isotropic nature of the data. In this case, one should be very cautious when attempting to use the hierarchical representations encoded in the network since neural networks are fundamentally interpolation methods \cite{mallat2016understanding} and lacking of interpretations often suggests the representations are hard to generalize.

\subsubsection{Sea surface temperature}
In this example, we consider the global sea surface temperature (SST) data. The NOAA OI SST V2 data set is open source and can be downloaded at \url{http://www.esrl.noaa.gov/psd/}. The data is collected each month and spans a 20 year period from 1990 to 2010.
\begin{figure}[t]
    \centering
    \label{fig:sst}\includegraphics[width=130mm, height=150mm]{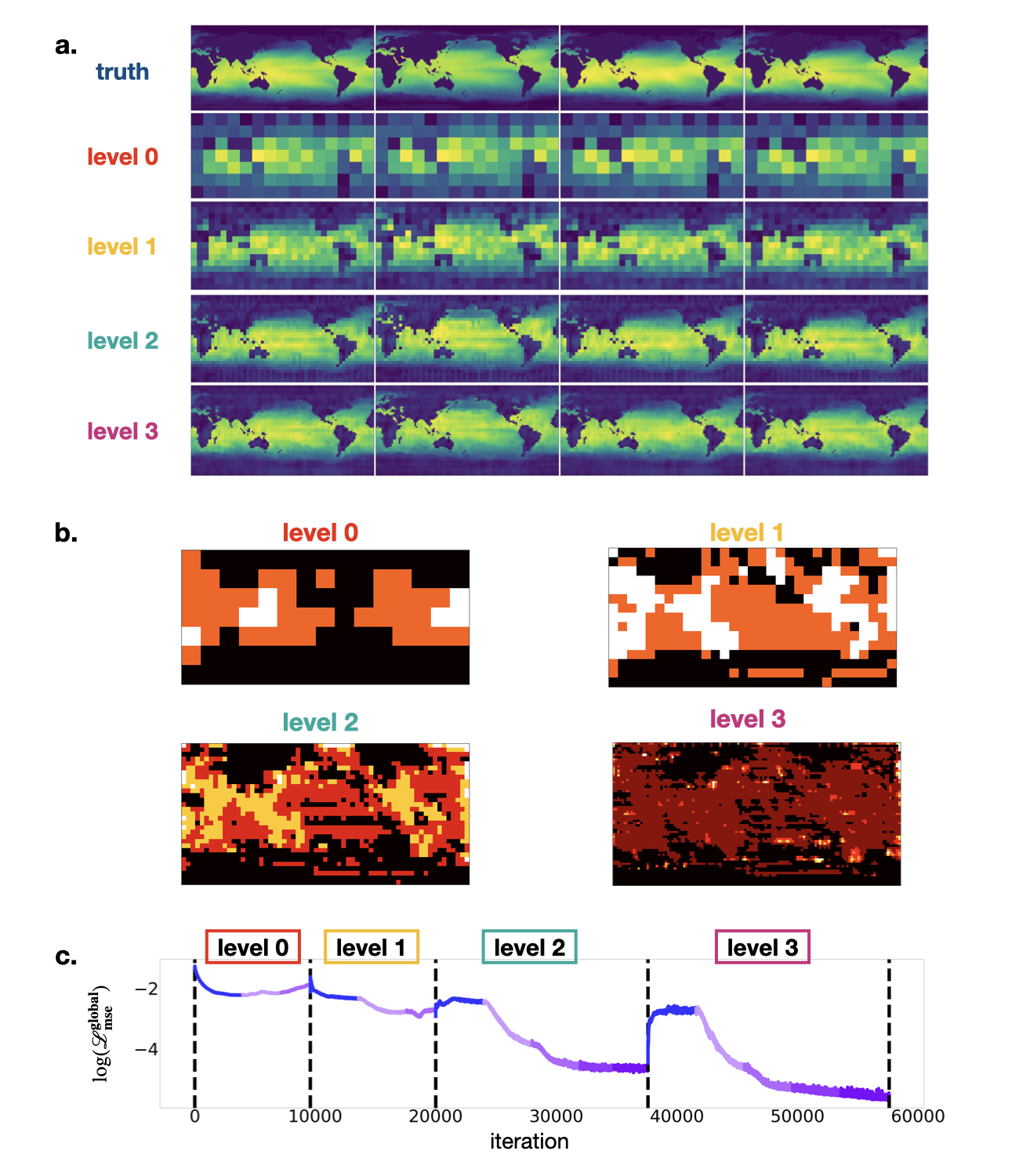}
    \vspace*{-0.4in}
    \caption{\textbf{a}. Original data and reconstructions of the sea surface temperature example across different levels of the network over the sampled test snapshots. \textbf{b}. Heat maps that reflect the regions that adaptive filters apply across different levels of the architecture. The brighter the pixel, the more filters are used. \textbf{c}. Logarithmic error plot (in terms of the metric presented in \cref{sec:err}) on the validation set throughout the training.}
\end{figure}
In this example, we set up a network of $4$ levels with each level having $2$, $2$, $4$, $4$ groups of adaptive filters. \cref{fig:sst} shows the increasingly refined reconstructions at higher and higher levels and the heat maps of regions where the adaptive filters apply. One observation is that network explores the middle regions more exhaustively. We conjecture that the underlying reason could be the temperature at the North Pole (top) and the South Pole (bottom) are less variant. In addition, one can see the boundaries of the highly intensive regions align well with edges of the continents which suggests the transition from the land to the ocean.

\subsection{Benchmarks}
We benchmark MrCAE against the RED-Net, which is structurally similar to ours. We run experiments of the network with and without rectified nonlinear units (ReLU) which results in two candidate comparisons: RED-Net(Linear) and RED-Net(+ReLU). We also include our MrCAE without adaptive filters, but equipped with the rectified nonlinear units, which is named MrCAE(+ReLU). 
\par
Although RED-Nets were not primarily proposed to do progressive training with data of increasing spatio-temporal resolution, for fair comparison, we also use different resolution data to train them and use the metric in \cref{sec:err} to evaluate their performance. 
\par
The procedure is as follows: throughout the training of our MrCAE network, we obtain a sequence of network architectures at different hierarchical levels. For each specific architecture, we train a corresponding one (with the same network depth and the same number of filters across each layer) for all other networks used for comparison. The key differences among the four candidate networks are as follows:
\begin{itemize}
    \item \textbf{number of parameters}: MrCAEs have sparse connections which requires less parameters whereas connections in RED-Nets are dense. This is due to the fact that different group of filters are independently patched in MrCAEs.
    \item \textbf{size of encoding}: MrCAE(PR) explicitly utilizes a sparse coding scheme whereas the other three types of networks don't have this feature.
\end{itemize}
\par
Results are shown in \cref{fig:n_params} and \cref{fig:encodings}. Axes in both plots are in logarithmic scales.

\subsubsection{Number of parameters}

\begin{figure}[t]
    \centering
    \label{fig:n_params}\includegraphics[width=130mm, height=62mm]{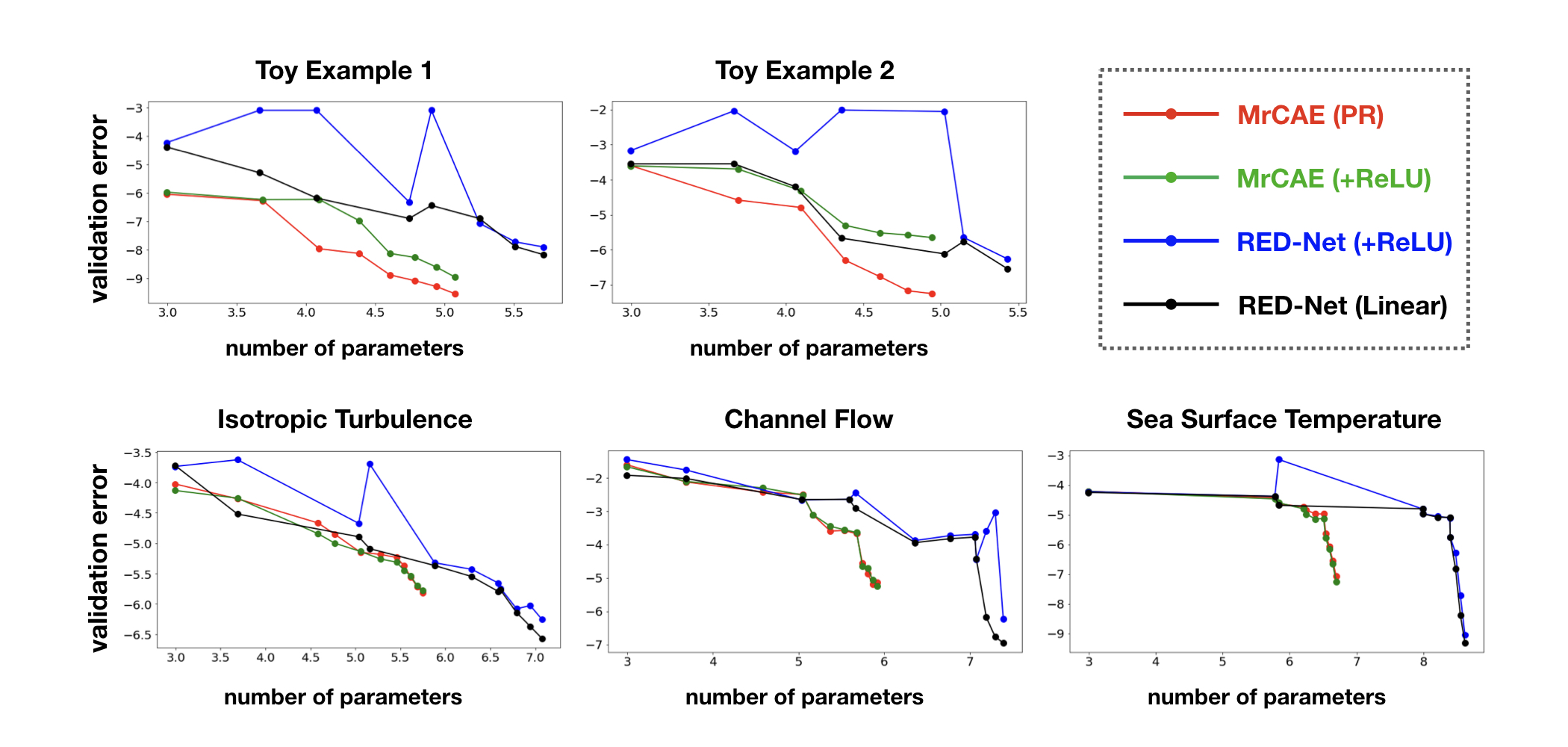}
    \vspace*{-.3in}
    \caption{\textbf{MrCAEs require less parameters.} 
    Comparison of the validation error versus number of parameters for our method versus other state-of-the-art methods.  Both the number of parameters and the validation error are on logarithmic scales. In this figure, each point represents a  particular architecture: we progressively train the networks of all kinds so that their depths and widths increase along the curves. The number of parameters of MrCAEs scale better as the network expands while exhibiting a steady decrease in validation error.}
\end{figure}

In \cref{fig:n_params}, we show that MrCAEs use significantly fewer parameters in comparison with similar architectures of RED-Nets. In other words, due to the sparse connections, it scales better when the network grows. Moreover, the error curves for MrCAEs seem to steadily decrease as the network grows which is highly desirable. But for RED-Nets, the curves are quite variable --- more parameters does not necessarily offer better results. And despite the parsimonious use of parameters, MrCAEs can achieve reasonable accuracy and sometimes, in the toy examples where the dynamics are not that complex, outperform other networks. We conjecture the reasons are of two folds: First, this spatio-temporal data has a clear separation of scales and our adaptive filters are primarily designed to efficiently leverage this feature. And second, the progressive training process may have some guiding effects for the network parameters which lowers the risks of getting stuck in some local minima. 

\subsubsection{Size of encoding}

\begin{figure}[t]
    \centering
    \label{fig:encodings}\includegraphics[width=130mm, height=62mm]{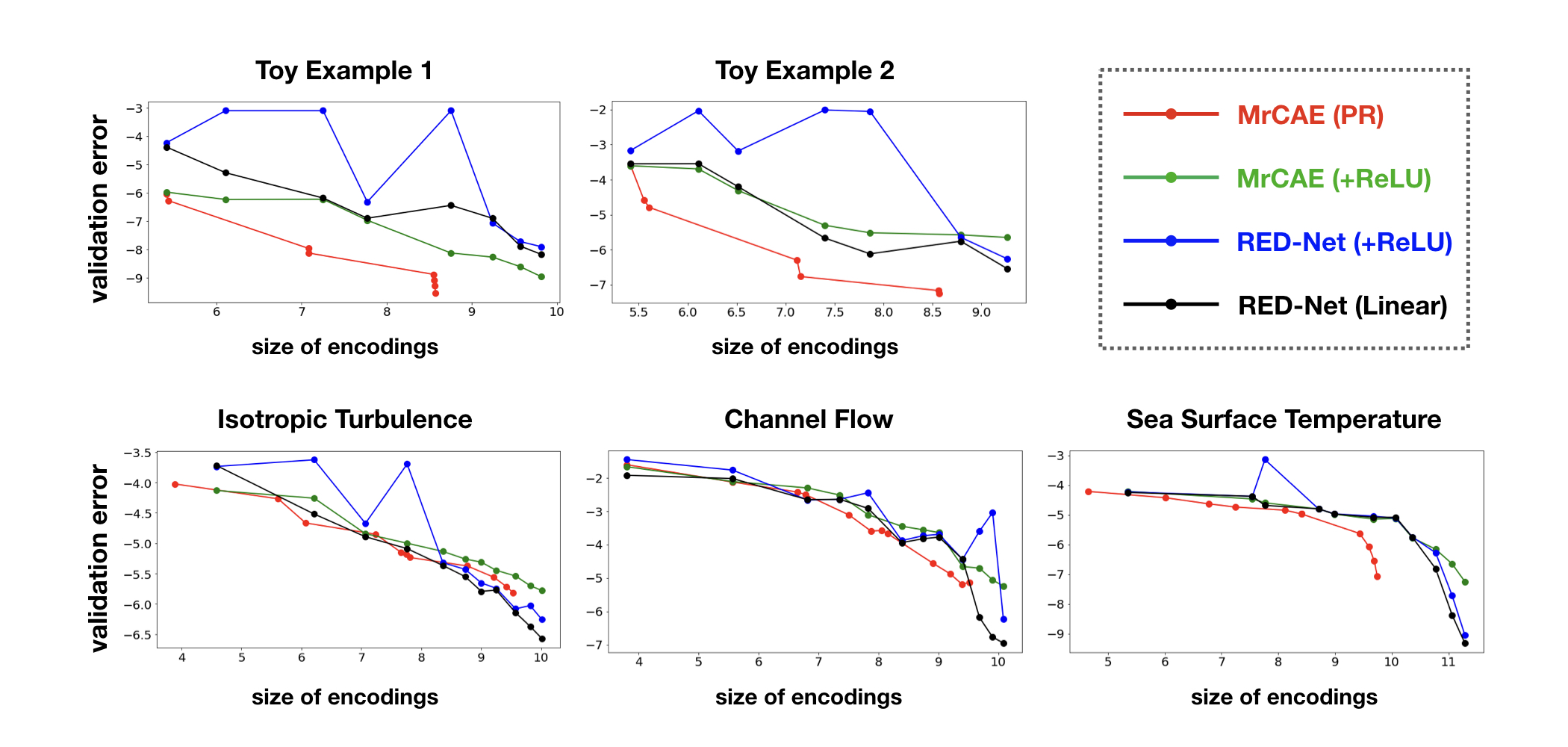}
    \vspace*{-.3in}
    \caption{\textbf{MrCAE(PR) offers better compression ratio}. Comparison of the validation error versus size of encoding for our method versus other state-of-the-art methods. Both the size of encoding and the validation error are on logarithmic scales. MrCAE provides smaller sizes of encoding over similar architectures due to the exploitation of spatial patterns in data sets.}
\end{figure}

In \cref{fig:encodings}, we show the validation error versus the size of encoding. By utilizing the adaptive filters, we are able to hierarchically encode different regions: some filters are only applied to highly variable regions. Therefore in practice, with similar architecture, MrCAE(PR) can always generate a smaller encoding size which leads to a better compression ratio. (It should be noted that unlike other types of networks, in principle, we should count not only the encoded information but also the corresponding positions which is similar to storing sparse matrices when we calculate size of encoding for MrCAE. However, since the position information is shared across all the snapshots and in practice there would be a large number of snapshots, the size of the position information becomes negligible, so we don't include it in our calculation.)
In the isotropic turbulence example, this gain is less obvious because of its isotropic nature: there's very little structured spatio-temporal data in this application.

\section{Discussion}
\label{sec:discussions}
In summary, we have equipped the highly-successful  convolutional autoencoder (CAE) with both adaptive filters and multiscale modeling capabilities, giving rise to a flexible workflow which allows for improved interpretability , control of the modeling framework, and in-depth understanding over modern end-to-end architectures.
Our proposed MrCAE architecture integrates and leverages three highly successful mathematical architectures: (i) multigrid methods, (ii) convolutional autoencoders and (iii) transfer learning.
Instead of training an extremely large black box model end-to-end, our model progressively utilizes a refinement strategy to build a hierarchical structure which leverages increased data due to improved spatial resolution. As a consequence, it enables automatic data augmentation across different spatial resolutions, and outputs insightful intermediate models and results for distinct spatial scales. These intermediate models can either guide the next-level architecture design or be put into immediate use which may be favored by online algorithms. 
\par
In addition, we develop a masking mechanism for the use of adaptive convolutional filters which resembles the mesh refinement process. It fully exploits the spatial patterns (which often shows up in many interested spatio-temporal physical systems) in the data set and therefore the size of encoding is small, the use of parameters is parsimonious in nature and the interpretation for each convolutional filter is clear. Though it is highly effective in resolving reconstruction errors, cautions must be taken when there is no clear spatio-temporal separation of scales in the data set. 
\par
There are many ongoing challenges and promising directions that motivate future works. Our network is currently very shallow. However, many successful applications achieved by neural networks rely heavily on its depth, as these architectures are reported to create more abstractions and therefore finding more efficient representations \cite{goodfellow2016deep}. Moreover, our original objective for building up this framework is to do multi-scale forecasting. By adopting such an encoder-decoder architecture, now we are able to build different forecasting tools for different parts of the hierarchical representations. Then, through the decoder, we are able to map it back to the original space. It will be very interesting to see how different forecasting algorithms will be built into the network and how they will affect the encoder-decoder architecture.

\appendix
\section{Remarks on filter initialization}
\label{sec:filter_init}
As we perform the deepening operation or widening operation, we are essentially adding properly initialized convoltuional and deconvolutional filters to the existing network.
\par
For a deepening operation, one convolutional filter $C_0$ and one deconvolutional filter $D_0$ are added to the network to connect the inputs and outputs from two adjacent levels. Both of the two filters are of kernel size $3 \times 3$ and applied with stride size $2$ without padding, and they are initialized with the following matrices: 

\begin{equation}
    C_0 = \begin{bmatrix}
    0 & 0 & 0 \\
    0 & 1 & 0 \\
    0 & 0 & 0 
    \end{bmatrix} 
    + \begin{bmatrix}
    \epsilon_{11}^{(c)}  & \epsilon_{12}^{(c)}  & \epsilon_{13}^{(c)}  \\
    \epsilon_{21}^{(c)}   & \epsilon_{22}^{(c)}  & \epsilon_{23}^{(c)}  \\
    \epsilon_{31}^{(c)}   & \epsilon_{32}^{(c)}  & \epsilon_{33}^{(c)} 
    \end{bmatrix}
\end{equation}
\begin{equation}
    D_0 = \begin{bmatrix}
    \frac{1}{4}  & \frac{1}{2} & \frac{1}{4} \\
    \frac{1}{2} & 1 & \frac{1}{2} \\
    \frac{1}{4} & \frac{1}{2} & \frac{1}{4}
    \end{bmatrix} 
    + \begin{bmatrix}
    \epsilon_{11}^{(d)}  & \epsilon_{12}^{(d)} & \epsilon_{13}^{(d)} \\
    \epsilon_{21}^{(d)}  & \epsilon_{22}^{(d)} & \epsilon_{23}^{(d)} \\
    \epsilon_{31}^{(d)}  & \epsilon_{32}^{(d)} & \epsilon_{33}^{(d)}
    \end{bmatrix}
\end{equation}
In a nutshell, $C_0$ is initialized with a down-sampling operator while $D_0$ is initialized with an interpolation operator so that the transition is smooth between each level. We also add some uniformly distributed random noise ($\epsilon_{ij}^{(c)}, \epsilon_{ij}^{(d)} \in [-\epsilon, \epsilon]$) to break the symmetry.   
\par 
To perform a widening operation, we create a group of convolutional filters and deconvolutional filters where the channel number is specified by the user. Also for the sake of ensuring a smooth transition, we initialize all entries of these filters with independent, uniformly distributed random variables from $[-\epsilon, \epsilon]$ so that immediately after the change of the architecture, reconstructions won't be affected by much. 

\section{Remarks on training}
\label{sec:early_stop}
In addition to specifying the tolerances (threshold for residuals on all pixels) for different training phases as shown in \cref{alg:pt}, we set a maximum number of epochs to stop the training. We also implement a stopping criterion as the convergence slowns down: if the relative error decrease every ten epochs is less than $0.001$, we stop the training and move to the next operation to expand the network capacity.

\section*{Acknowledgements}
This work was performed under the auspices of the U.S. Department of Energy by Lawrence Livermore National Laboratory under Contract DE-AC52-07NA27344 and was supported by the LLNL-LDRD Program under Project No. 19-ERD-019, LLNL-JRNL-808088-DRAFT.  SLB acknowledges support from the Army Research Office (ARO W911NF-17-1 0306). 
JNK acknowledges support from the Air Force Office of Scientific Research (AFOSR) grant FA9550-17-1-0329.   

This document was prepared as an account of work sponsored by an agency of the United States government. Neither the United States government nor Lawrence Livermore National Security, LLC, nor any of their employees makes any warranty, expressed or implied, or assumes any legal liability or responsibility for the accuracy, completeness, or usefulness of any information, apparatus, product, or process disclosed, or represents that its use would not infringe privately owned rights. Reference herein to any specific commercial product, process, or service by trade name, trademark, manufacturer, or otherwise does not necessarily constitute or imply its endorsement, recommendation, or favoring by the United States government or Lawrence Livermore National Security, LLC. The views and opinions of authors expressed herein do not necessarily state or reflect those of the United States government or Lawrence Livermore National Security, LLC, and shall not be used for advertising or product endorsement  purposes.

\newpage
\bibliographystyle{siamplain}
\bibliography{references}

\end{document}